\documentclass[twoside,11pt]{article}

\usepackage{jmlr2e, amsmath, footmisc}

\ShortHeadings{Dynamic Mortality Risk Prediction}{Aczon, Ledbetter, Ho, et. al}
\firstpageno{1}

\begin{document}

\title{Dynamic Mortality Risk Predictions in Pediatric Critical Care Using Recurrent Neural Networks}
\author{}
\maketitle

{\centering{\bf{Melissa Aczon\footnote[2]{Correspondence: \{maczon, dledbetter, loho, agunny, aflynn, jonwilliams, rwetzel\}@chla.usc.edu}}, David Ledbetter\footnote[1]{The first two authors contributed equally to this work.}, Long Van Ho, Alec Gunny, \\
			Alysia Flynn, Jon Williams, Randall Wetzel} \\[2pt]
			The Laura P. and Leland K. Whittier Virtual Pediatric Intensive Care Unit\\
			Children's Hospital Los Angeles \\[25pt]} 

\begin{abstract}
Viewing the trajectory of a patient as a dynamical system, a recurrent neural network was developed to learn the course of patient encounters in the Pediatric Intensive Care Unit (PICU) of a major tertiary care center.  Data extracted from Electronic Medical Records (EMR) of about 12000 patients who were admitted to the PICU over a period of more than 10 years were leveraged.  The RNN model ingests a sequence of measurements which include physiologic observations, laboratory results, administered drugs and interventions, and generates temporally dynamic predictions for in-ICU mortality at user-specified times.  The RNN's ICU mortality predictions offer significant improvements over those from two clinically-used scores and static machine learning algorithms.
\end{abstract}

\section{Introduction}

\subsection{Background on severity and mortality scores}
Numerous severity of illness (SOI) and mortality scoring systems have been developed over the past three decades [\cite{le2005use, strand2008severity}].  Two of the earliest and commonly used scores are Acute Physiology and Chronic Health Evaluation (APACHE II), [\cite{knaus1985apache}] and Simplified Acute Physiology Score (SAPS) [\cite{le1984simplified}], both of which rely on routine physiologic measurements and the deviations of those measurements from expert-defined {\it{normal}} values.  In pediatric critical care, the Pediatric Risk of Mortality (PRISM), [\cite{pollack1988pediatric}] score and Pediatric Index of Mortality (PIM), [\cite{shann1997paediatric}] were developed.  Both leverage physiologic data, with PIM incorporating into its calculations additional information such as pre-ICU procedures and in-ICU ventilation data from the first hour.  Larger databases led to refinements of these systems; for some examples, see APACHE III in \cite{knaus1991apache}; SAPS 3 in \cite{moreno2005saps}; PRISM 3 in \cite{pollack1996prism}; PRISM 4 in \cite{pollack2016pediatric}; PIM 2 in \cite{slater2003pim2}; PIM 3 in \cite{straney2013paediatric}. \cite{pollack2016severity} makes a distinction between scoring for severity of illness or predicting mortality.  Regardless of this distinction, however, the effectiveness of these models are measured via their ability to discriminate between surviving and non-surviving patients.

\subsection{EMR and advanced computing methods}
The adoption of Electronic Medical Records (EMR) has enabled ready access to more variables and more patients.  In response to this ever-growing amount of data, machine learning techniques increasingly have been used to develop models which forecast patient condition.  The Gaussian process-based scores in \cite{ghassemi2015multivariate} and \cite{alaa2016personalized} leverage time-series measurements instead of static values from a fixed time window of the systems described earlier.  Towards personalized scoring, the latter attempts to account for heterogeneity by discovering, via unsupervised learning, some number of {\it{classes}} in the population, then learns the parameters that govern each class. The Rothman index [\cite{rothman2013development}] also offers continuously updated scores but still generates predictions using a static snapshot. 

The use of neural networks in ICU applications actually goes back more than 20 years, as reviewed in \cite{hanson2001artificial} . In general those studies were relatively small in scale (hundreds of patients). Since then, two primary factors have changed the landscape.  First, larger datasets containing tens of thousands of patients with millions of measurements are now available. Second, computing hardware advances in the last decade, particularly Graphics Processing Units (GPUs), have enabled larger, deeper networks to be trained.  These more sophisticated networks have demonstrated remarkable success in wide-ranging applications such as computer vision [\cite{krizhevsky2012imagenet,he2015deep}], speech recognition [\cite{hinton2012deep}], and natural language processing [\cite{mesnil2015using}].

\subsection{Summary of RNN-based mortality model/framework}

Recurrent Neural Networks (RNN) were designed to process sequential data.  A key feature is a {\it{feedback loop}} which allows integration of information from previous steps with newly acquired data.  Thus, they provide an elegant infrastructure to process ever-evolving streams of clinical data.  Figure 1 gives a high-level illustration of data ingestion and prediction with a RNN, and Section 3.1 will describe the mathematical formulation.  Each input vector ($x$) contains clinical measurements (e.g. physiology and medications).  The infrastructure also allows the user to specify how long into the future predictions are for.  Each output vector ($y$) contains forecasts such as mortality risk at the specified future time.  This work focused on predicting in-ICU mortality of pediatric patients.  Hospital mortality is used because it is an objective function to assess performance despite its inability to  capture more subjective components of care such as quality of survival [\cite{knaus1985apache}].  Also, the sequential nature of the RNN's mortality risk predictions provides a dynamic tracking of patient condition.  


\begin{figure}[htbp]
  \centering 
  \includegraphics[scale=0.45]{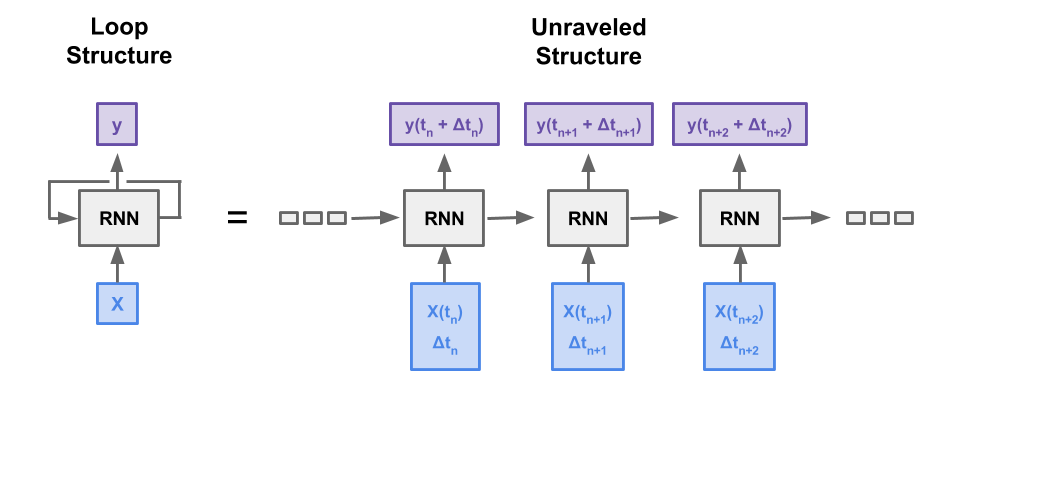} 
  \caption{Flow of data in RNN-based framework.  When measurements (vector $X$) become available at time $t_n$, they are ingested as inputs to the RNN kernel.  The RNN then generates a prediction $Y$ corresponding to a future time specified by the user through $\Delta t_n$.} 
  \label{fig:rnn_overview1} 
\end{figure} 


\section{Data and pre-processing}

 We leveraged anonymized EMR from the PICU at Children's Hospital Los Angeles between December 2002 and March 2016. The data for each patient included static information such as demographics, diagnoses, and disposition (alive or not) at the end of the ICU encounter.  An encounter is defined as a contiguous admission into the PICU.  Each encounter contained irregularly, sparsely and asynchronously sampled measurements of physiologic observations (e.g. heart rate, blood pressure), laboratory results (e.g. creatine, glucose level), drugs (e.g. epinephrine, furosemide) and interventions (e.g. intubation, oxygen level).  

A single patient can have multiple encounters, and this is an important point for validation.  When the database was split into training and testing sets, all encounters from a single patient belonged to exactly one of these sets to prevent possible leakage.  Encounters that did not include disposition information were excluded from the final database used for the results presented here.  This database consisted of 12020 patients with 16559 encounters. Seventy-five percent of the patients were randomly selected and placed into the training set, and the remaining twenty-five percent into a holdout set for testing.  This splitting resulted in 12460 encounters (with 4.85\% mortality) in the training set, and 4099 encounters (with 5.12\% mortality) in the holdout set.

\begin{figure}[htbp]
  \centering 
  \includegraphics[scale=0.8]{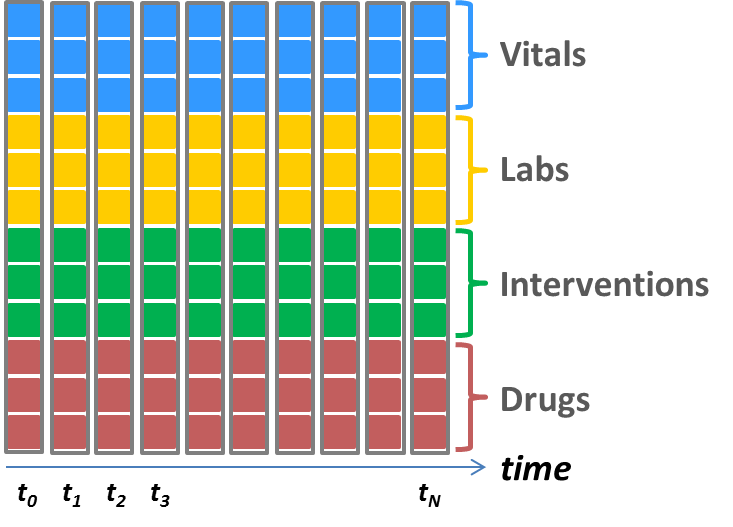}
  \caption{After pre-processing, data for a single patient encounter is in a matrix format.  A single row of data contains values (actual and imputed measurements) from a single variable. A column of data comprises all measurements at one time point.}
  \label{fig:data_schema} 
\end{figure}

To leverage existing deep learning frameworks, the data were first converted into the matrix format illustrated in Figure \ref{fig:data_schema} with a pre-processing pipeline described below.

\subsection{Constituent Aggregation and Normalization}

Similar physiologic observations or laboratory measurements were aggregated into a single variable.  For example, 
non-invasive and invasive measurements of systolic blood pressures were grouped together into a single systolic blood pressure variable.  This aggregation resulted in approximately 300 different physiologic and treatment variables, a complete list of which can be found in Appendix A.

All quantities under the same variable were converted into the same unit of measure.  Drugs and some interventions were converted to a binary variables corresponding to absence or presence of administration.  Variables that were not binarized were Z-normalized.  The mean and standard deviation needed for this transformation were computed from the training set only, and then applied to both the training and holdout sets.

\subsection{Imputation}

The measurements in the database were {\it{sparse}}, {\it{irregularly sampled}} -- time between any two consecutive time points ranges from a minute to several hours -- and {\it{asynchronous}}.  At any time point when at least one variable had a recorded value, the values for all other variables without a measurement at that point were imputed.  The imputation process depended on the variable type.  Any missing measurement of a drug or an intervention variable was imputed as zero.  When a physiologic observation or lab measurement was available, it was propagated forward until its next reading.  However, if a physiologic or laboratory variable had no recorded value throughout the entire encounter, then that variable was set to zero at all time points for that encounter.  Since physiologic observations were first Z-normalized, a zero imputation is equivalent to an imputation with the population mean derived from the training set. Note that no features were age-normalized; instead age was an input as a physiologic observation. These choices were based on a reasonable assumption about clinical practice:  measurements are taken more frequently when {\it{something}} is happening to the patient, and less frequently when the patient appears {\it{stable}}.


\section{RNN-based Framework}\label{sec:rnn}

\subsection{From dynamical system to RNN}
\begin{figure}[htbp]
	\centering 
	\includegraphics[scale=0.45]{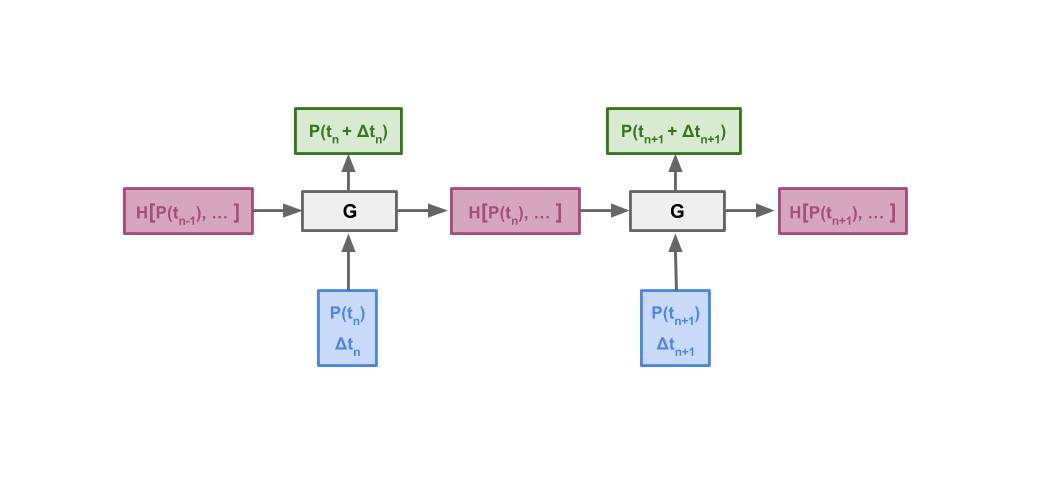} 
	\caption{Visual diagram of finite difference formulation given in Equation \ref{eq:3}.}
	\label{fig:funcG_funcH} 
\end{figure}

The trajectory of a patient can be viewed as a continuous dynamical system composed of many variables, $P(t) = [{\rm {vitals, labs, drugs, interventions}}]^T$, interacting with each other:
\begin{equation}
\frac{dP(t)}{dt} = F\bigl[ P(t) \bigr], \qquad P(t_0) = P_0. \label{eq:1}
\end{equation}
{\noindent In equation (\ref{eq:1}), $F$ denotes the unknown and complex function governing the variable interactions, while $P_0$ is the state at some initial time, $t_0$.  For the PICU setting and data, $t_0$ corresponds to the start time of an ICU encounter.  
	
	Finite difference methods are a standard way to solve equation (\ref{eq:1}) [\cite{leveque2007finite}].  Any such approximation can be cast into the form}
\begin{equation}
P(t_{n} + \Delta t_n) = G\bigl[P(t_n), \Delta t_n, P(t_{n-1}), \ldots, P(t_{n-k}), \bigr], \label{eq:2}
\end{equation}
where $t_n + \Delta t_n$ is a specified future time of interest.  This can be further abstracted into the form:
\begin{equation}
P(t_{n} + \Delta t_n) = G\biggl[P(t_n), \Delta t_n, H\bigl[P(t_{n-1}), \ldots, P(t_{n-k})\bigl] \biggr], \label{eq:3}
\end{equation}
where $H$ is a transformation of data from previous time steps.  Equation (\ref{eq:3}) -- which can be visualized in Figure \ref{fig:funcG_funcH} -- is a recurrent relation, and $G$ describes the mapping from past states into a future state.  

\begin{figure}[htbp]
	\centering 
	\includegraphics[scale=0.45]{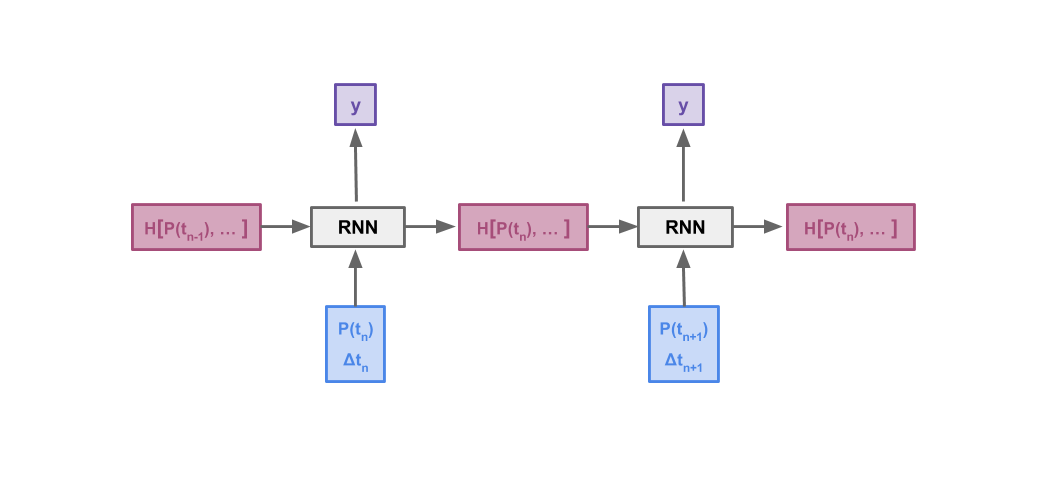} 
	\caption{The output of the RNN, $y$, is another transformation }
	\label{fig:ypred_RNN} 
\end{figure}

Independent work by \cite{funahashi1989approximate} and \cite{hornik1989multilayer} established that any function with mathematically reasonable properties can be approximated by a neural network to an arbitrary degree of accuracy, i.e.
\begin{equation}
G = \underbrace{\sum_i \alpha_i \sigma\biggl[W_i \biggl(P(t_n), \Delta t_n, 
	H\bigl[P(t_{n-1}), \ldots, P(t_{n-k})\bigl] \biggr) + b_i \biggr]}_{\text{Neural Network:}\ N} +\ \epsilon, \label{eq:4}
\end{equation}
where $\epsilon$ is an arbitrarily small real number.  The finite difference formulation therefore becomes
\begin{equation}
P(t_{n} + \Delta t_n)\ \approx\ N \biggl[P(t_n), \Delta t_n, H\bigl[P(t_{n-1}), \ldots, P(t_{n-k})\bigl] \biggr]. \label{eq:5}
\end{equation}
The output of $H$ can be regarded as a {\it{hidden state}} or an internal representation of the patient's history, and the mortality risk can be inferred from this integration of history.  The diagram of Figure \ref{fig:funcG_funcH} then leads to the RNN formulation shown in Figure \ref{fig:ypred_RNN}, where the RNN module encapsulates $G$ from Equation (\ref{eq:2}) and an additional function that transforms the internal state to some observable manifestation, such as mortality risk, represented by the output $y$.  The recurrent aspect of the network, i.e. the feedback mechanism, allows past information to be propagated forward.

It is worth noting that work by \cite{funahashi1993approximation} showed a direct path from dynamical systems to a class of continuous time recurrent neural networks (CTRNN) provided the original function, $F$, in Equation (\ref{eq:1}) meets continuity conditions.  \cite{chow2000modeling} and \cite{li2005approximation} extended the theory to handle more general dynamical systems, including time-variant ones with inputs for control .  Applying a finite difference scheme to the CTRNN leads to a form that is very similar to Equation (\ref{eq:5}).

\begin{figure}[htbp]
  \centering 
  \includegraphics[scale=0.45]{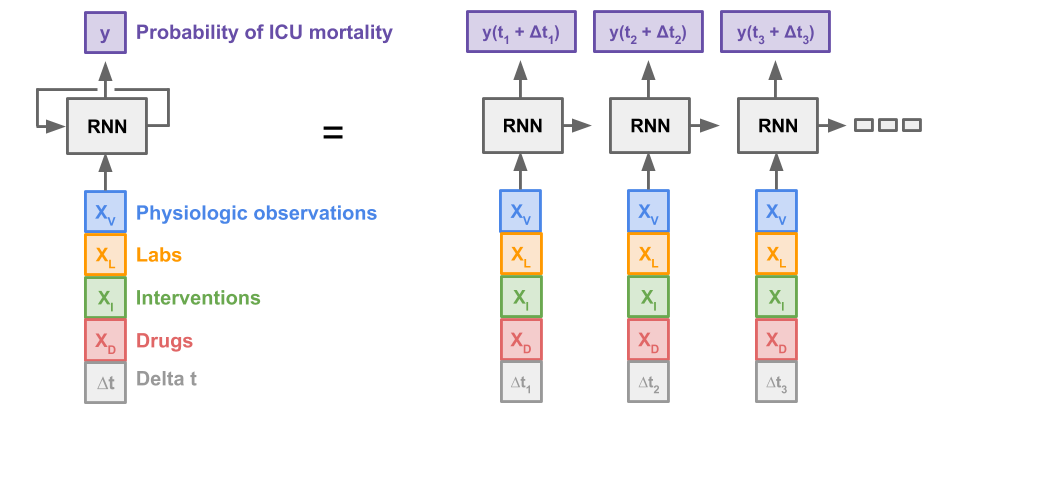} 
  \caption{Physiologic observations, laboratory measurements, interventions and drugs at time $t_n$ are inputs to the RNN kernel.  The RNN then projects a mortality risk for time $t_n + \Delta t_n$, where $\Delta t_n$ is specified by a user.}
  \label{fig:rnn_picu} 
\end{figure} 

\subsection{RNN architecture and implementation}

Figure \ref{fig:rnn_picu} illustrates the flow of the PICU data and predictions in the RNN-based infrastructure.  The input vector to the network at time $t_n$ consists of five main groups of measurements from a patient's ICU encounter: $\big[X_V(t_n), X_L(t_n), X_I(t_n), X_D(t_n), \Delta t_n\big]^T$. The vector $X_V$ contains the physiologic observations, the vector $X_L$ contains laboratory results, the vector $X_I$ comprises the interventions, $X_D$ records the administered drugs, and the scalar $\Delta t_n$ specifies how far into the future the user wants to forecast.  Including $\Delta t_n$ in the input vector follows naturally from the finite difference formulation (\ref{eq:2}) and serves a dual purpose: it gives the user control and flexibility in time-to-prediction, and it also enables augmentation of the training data during model development.  The output at this time step is a probability of survival at the future time $t_n + \Delta t_n$ which can be thought of as a prediction of patient condition at that future time. 

\begin{figure}[htbp]
	\centering 
	\includegraphics[scale=0.5]{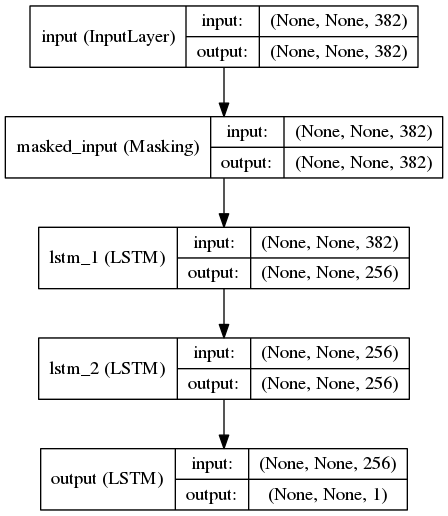} 
	\caption{RNN architecture for PICU data}
	\label{fig:rnn_model} 
\end{figure}

A number of RNN architectures have been developed and studied \citep{greff2015lstm, jozefowicz2015empirical}.  The specific one utilized here is the Long Short-Term Memory (LSTM) architecture of \cite{hochreiter1997long}.  The Keras python deep learning framework \citep{keras2015} was used to construct a model comprised of three LSTMs -- see Figure \ref{fig:rnn_model} -- and train this model to make predictions for in-ICU mortality.

\section{Results} 

\begin{figure}[htbp]
	\centering 
	\includegraphics[scale=0.5]{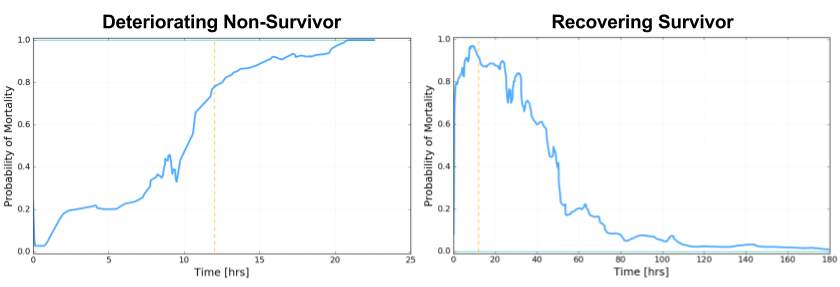} 
	\caption{RNN-generated mortality risk of two patients tracked over their ICU encounter.  The dashed yellow lines indicate the 12-hour mark.  The patient on the left slowly deteriorated over the course of a day and did not survive.  The patient on the right started as very high-risk but recovered over the course of a week.}
	\label{fig:PatientsThroughTime_1} 
\end{figure}

The RNN model continuously updates its mortality risk predictions as it intakes new data. Figure \ref{fig:PatientsThroughTime_1} displays these temporally evolving risk scores from two patients.  This dynamic tracking, which is automatic in the RNN, is not done by PIM 2 or PRISM 3.  The Rothman index updates its predictions when new measurements become available, but its update does not integrate past measurements.  In this sense, the Rothman index still processes time series data in a static manner, while the RNN dynamically integrates data through its feedback mechanism.    

The RNN's ICU mortality predictions were compared to those of PIM 2 and PRISM 3, both of which were pulled directly from the EMR.  A logistic regression (LR) and a multi-layer perceptron (MLP) were also implemented for additional comparisons.  The LR, MLP and RNN models access identical clinical data.  Like PIM 2 and PRISM 3, LR and MLP are {\it{static}} methods, meaning they process a snapshot of data from a fixed window of time to make a single-time prediction.  Again, this is a contrast to the RNN which continuously integrates incoming data with past information.  Both PIM 2 and PRISM 3 use information collected prior to ICU admission, data which the RNN, MLP and LR models do not access.  In addition, PIM 2 incorporates some data from the first hour in the ICU, while PRISM 3 uses data from the first 12 hours in the ICU. 

\begin{figure}[htbp]
  \centering 
  \includegraphics[scale=0.4]{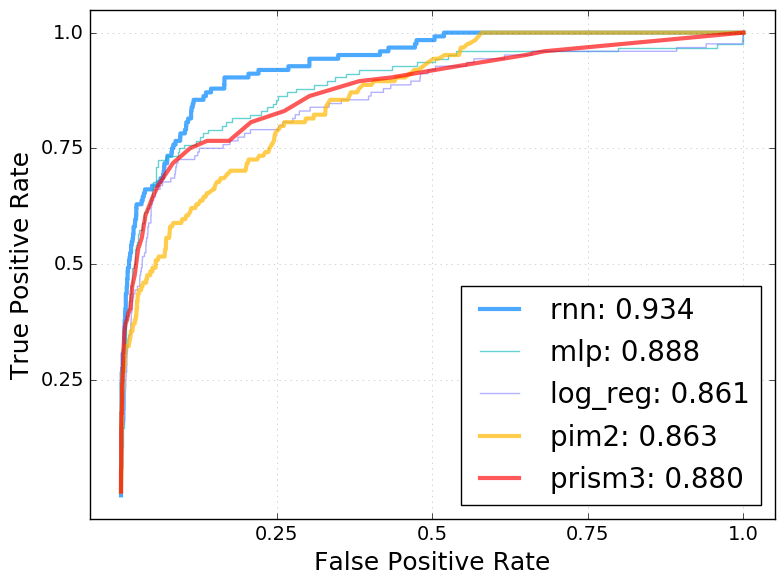} 
  \caption{Comparison of ICU mortality predictions from various models:  Recurrent Neural Network, Multi-Layer Perceptron, Logistic Regression, PIM2, and PRISM3.  ROC curves and AUCs were generated from 2849 holdout encounters that had both PIM 2 and PRISM 3 scores available.}  \label{fig:MultiAlgs_ROCs} 
\end{figure}

The ICU mortality predictions of the different models were assessed via Receiver Operating Characteristic (ROC) curves and corresponding Area Under the Curve (AUC).  Figure \ref{fig:MultiAlgs_ROCs} shows the results on $2849$ holdout encounters that had both PIM 2 and PRISM 3 scores.  The RNN model yields an AUC of $93.4\%$ which is significantly higher than the comparative models [MLP: $88.8\%$ ($p < 0.01$), LR: $86.1\%$ ($p < 0.001$), PIM 2: $86.3\%$ ($p < 0.001$), and PRISM 3: $88.0\%$ ($p < 0.003$)].  The LR, MLP and RNN predictions used to generate these results were all taken from the 12th hour.  The difference in performance between the MLP and RNN provides a rough quantification of the boost that dynamic integration provides over static computation.  The feedback loop gives the RNN a temporal memory which enables temporal trends -- i.e., function derivatives -- to be incorporated into the model.  

\begin{figure}[htbp]
  \centering 
  \includegraphics[scale=0.4]{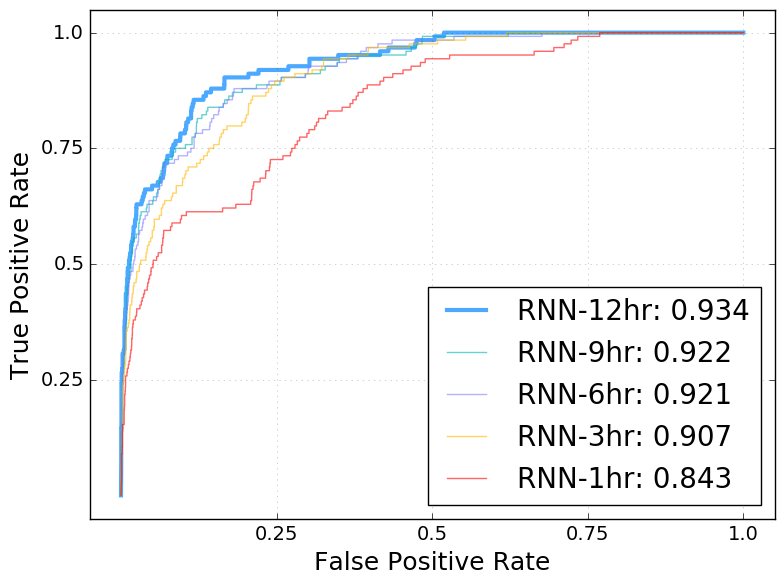} 
  \caption{Comparison of ICU mortality predictions from RNN model after various lengths of observation time.  These results were aggregated from the same 2849 holdout encounters in the previous figure.  All had least 12 hours of data.}  \label{fig:RNN_Nhrs} 
\end{figure} 

Figure \ref{fig:RNN_Nhrs} demonstrates the improved predictive capability of the RNN model as a function of increasing observation time.  After only three hours of observation, the RNN's AUC surpasses that of PRISM 3 which incorporates 12 hours of observation. As the RNN's observation window increases, the accuracy of its prediction continues to increase.  This is a desirable characteristic of a risk score. 


\section{Conclusions}

Recurrent Neural Networks were applied on ICU EMR to generate in-ICU mortality risk scores.  In addition to providing dynamic tracking of patient condition, the RNN-generated scores also achieved significantly higher accuracy [AUROC greater than 93\%] than the clinically used systems PIM 2 and PRISM 3.  The RNN model also outperformed logistic regression and multi-layer perceptron models.  The increased performance of the RNN model stemmed from two key factors:  access to more variables that characterize a patient and dynamic integration that allows it to incorporate temporal trends of those variables.  

Although approximately 300 variables have been encoded into the model, other data which are available from the PICU, such as fluid balance, have not been incorporated.  Monitor data, which have higher temporal resolution, also have yet to be included.  Future work will focus on aggregating these additional data and quantifying their impact on predictive accuracy.

\acks{This work was funded by a grant from the Laura P. and Leland K. Whittier Foundation.}

\bibliography{drtedv2}

\newpage
\appendix
\section*{Appendix A.  Clinical Data Used}

\begin{center}
\begin{tabular}{ l | l }
\multicolumn{2}{c} {Vitals} \\ \hline
Abdominal girth & Bladder pressure \\
Capillary refill rate & Central Venous Pressure \\
Cerebral perfusion pressure & Diastolic Blood Pressure \\
EtCO2 & Eye Response \\
GlascowCS & Head circumference  \\
Heart rate  & Height  \\
Intracranial pressure  & Left pupillary response \\
Motor Response & Near-infrared spectroscopy \% \\
Pulse Oximetry & Pupillary response \\
Respiratory rate  & Right pupillary response \\
Systolic Blood Pressure & Temperature  \\
Verbal Response & Weight \\
\end{tabular}
\end{center}

\begin{center}
\begin{tabular}{ l | l }
\multicolumn{2}{c} {Labs} \\ \hline
ABG Base excess  & ABG FiO2 \\
ABG HCO3  & ABG O2 sat \% \\
ABG PCO2  & ABG PO2  \\
ABG TCO2  & ABG pH \\
ALT (SGPT) & AST (SGOT) \\
Albumin level  & Alkaline phosphatase  \\
Amylase  & B-type Natriuretic Peptide  \\
BUN  & Bands \% \\
Basophils \% & Bicarbonate serum  \\
Bilirubin conjugated  & Bilirubin total  \\
Bilirubin unconjugated  & Blasts \% \\
C-reactive protein  & CBG Base excess \\
CBG FiO2 & CBG HCO3  \\
CBG O2 sat \% & CBG PCO2  \\
CBG PO2  & CBG TCO2  \\
CBG pH & CSF Bands \% \\
CSF Lymphs \% & CSF RBC \\
CSF Segs \% & CSF WBC \\
CSF glucose  & CSF protein  \\
Calcium ionized  & Calcium total  \\
Chloride  & Complement C3 serum  \\
Complement C4 serum  & Creatinine  \\
Culture CSF & Culture blood \\
Culture fungus blood & Culture respiratory \\
Culture urine & Culture wound \\
ESR & Eosinophils \% \\
FDP Titer & Ferritin level  \\
Fibrinogen  & GGT  \\
Glucose  & Haptoglobin  \\
\end{tabular}
\end{center}

\begin{center}
\begin{tabular}{ l | l }
\multicolumn{2}{c} {Labs (cont.)} \\ \hline
Hema & Hemo \\
INR & Lactate  \\
Lactate Dehydrogenase blood  & Lactic Acid blood  \\
Lipase  & Lymphocyte \% \\
MCH  & MCHC \% \\
MCV  & MVBG Base excess  \\
MVBG FiO2 & MVBG HCO3  \\
MVBG O2 sat \% & MVBG PCO2  \\
MVBG PO2  & MVBG TCO2  \\
MVBG pH & Macrocytes \\
Magnesium level  & Metamyelocytes \% \\
Monocytes \% & Myelocytes \% \\
Neutrophils \% & Oxygentaion index \\
P/F ratio & PT \\
PTT & PaO2/FiO2 \\
Phosphorus level  & Platelet count  \\
Potassium & Protein total  \\
RBC blood  & RDW \% \\
Reticulocyte count \% & Schistocytes \\
Sodium & T4 free  \\
TSH  & Triglycerides  \\
VBG Base excess  & VBG FiO2 \\
VBG HCO3  & VBG O2 sat \% \\
VBG PCO2  & VBG PO2  \\
VBG TCO2  & VBG pH \\
Virus & White blood cell count  \\
\end{tabular}
\end{center}

\begin{center}
\begin{tabular}{ l | l }
\multicolumn{2}{c} {Interventions} \\ \hline
Abdominal X ray & Alprostadil \\
Amplitude  & CT abdomen \\
CT abdomen/pelvis & CT brain \\
CT chest & CT pelvis \\
Chest X ray & Chest/abd X ray \\
Continuous EEG & ECMO hours \\
ECMO type & EPAP  \\
FiO2 & Foley catheter \\
Frequency  & Hemofiltration/CRRT \\
IPAP  & Inspiratory time  \\
MAP  & MRI brain \\
Mean airway pressure  & NIV set rate  \\
Nitric Oxide & O2 Flow  \\
PEEP  & PEEP  \\
Peak Inspiratory Pressure  & Peritoneal dyalisis \\
Pressure support  & Tidal volume delivered  \\
\end{tabular}
\end{center}

\begin{center}
\begin{tabular}{ l | l }
\multicolumn{2}{c} {Interventions (cont.)} \\ \hline
Tidal volume expiratory  & Tidal volume inspiratory  \\
Tidal volume set  & Tracheostomy \\
Ventilator rate  & Ventricular assist device \\
Volume Tidal
\end{tabular}
\end{center}

\begin{center}
\begin{tabular}{ l | l }
\multicolumn{2}{c} {Drugs} \\ \hline
Acetaminophen & Acetaminophen/Codeine \\
Acetaminophen/Hydrocodone & Acetaminophen/Oxycodone \\
Acetazolamide & Acetylcysteine \\
Acyclovir & Albumin \\
Albuterol & Allopurinol \\
Alteplase & Amikacin \\
Aminocaproic Acid & Aminophylline \\
Amiodarone & Amlodipine \\
Amoxicillin & Amoxicillin/clavulanic acid \\
Amphotericin B & Amphotericin B Lipid Complex \\
Ampicillin & Ampicillin/Sulbactam \\
Aspirin & Atenolol \\
Atropine & Azathioprine \\
Azithromycin & Baclofen \\
Basiliximab & Budesonide \\
Bumetanide & Calcium Chloride \\
Calcium Glubionate & Calcium Gluconate \\
Captopril & Carbamazepine \\
Carvedilol & Caspofungin \\
Cefazolin & Cefepime \\
Cefotaxime & Cefoxitin \\
Ceftazidime & Ceftriaxone \\
Cefuroxime & Cephalexin \\
Chloral Hydrate & Chlorothiazide \\
Ciprofloxacin HCL & Cisatracurium \\
Clarithromycin & Clindamycin \\
Clonazepam & Clonidine HCl \\
Clotrimazole & Cromolyn Sodium \\
Cyclophosphamide & Cyclosporine \\
Dantrolene Sodium & Desmopressin \\
Dexamethasone & Dexmedetomidine \\
Diazepam & Digoxin \\
Diphenhydramine HCl & Dobutamine \\
Dopamine & Dornase Alfa \\
Doxacurium Chloride & Doxorubicin \\
Doxycycline Hyclate & Enalapril \\
Enoxaparin & Epinephrine \\
Epoetin & Erythromycin \\
\end{tabular}
\end{center}

\begin{center}
\begin{tabular}{ l | l }
\multicolumn{2}{c} {Drugs (cont.)} \\ \hline
Esmolol Hydrochloride & Etomidate \\
Factor VII & Famotidine \\
Fentanyl & Ferrous Sulfate \\
Filgrastim & Flecainide Acetate \\
Fluconazole & Fluticasone \\
Fosphenytoin & Furosemide \\
Gabapentin & Ganciclovir Sodium \\
Gentamicin & Glycopyrrolate \\
Haloperidol & Heparin \\
Hydrocortisone & Hydromorphone \\
Ibuprofen & Imipenem \\
Immune Globulin & Insulin \\
Ipratropium Bromide & Isoniazid \\
Isoproterenol & Isradipine \\
Itraconazole & Ketamine \\
Ketorolac & Labetalol \\
Lactobacillus & Lansoprazole \\
Levalbuterol & Levetiracetam \\
Levocarnitine & Levofloxacin \\
Levothyroxine Sodium & Lidocaine \\
Linezolid & Lisinopril \\
Lorazepam & Magnesium Sulfate \\
Meropenem & Methadone \\
Methylprednisolone & Metoclopramide \\
Metolazone & Metronidazole \\
Micafungin & Midazolam HCl \\
Milrinone & Montelukast Sodium \\
Morphine & Mycophenolate Mofetl \\
Naloxone HCL & Naproxen \\
Nesiritide & Nifedipine \\
Nitrofurantoin & Nitroglycerine \\
Nitroprusside & Norepinephrine \\
Nystatin & Octreotide Acetate \\
Olanzapine & Ondansetron \\
Oseltamivir & Oxacillin \\
Oxcarbazepine & Oxycodone \\
Pancuronium & Pantoprazole \\
Penicillin G Sodium & Pentobarbital \\
Phenobarbital & Phenylephrine HCl \\
Phenytoin & Piperacillin \\
Piperacillin/Tazobactam & Potassium Chloride \\
Potassium Phosphate & Prednisolone \\
Prednisone & Procainamide \\
Propofol & Propranolol HCl \\
\end{tabular}
\end{center}

\begin{center}
\begin{tabular}{ l | l }
\multicolumn{2}{c} {Drugs (cont.)} \\ \hline
Prostacyclin & Protamine \\
Racemic Epi & Ranitidine \\
Rifampin & Risperidone \\
Rocuronium & Sildenafil \\
Sodium Bicarbonate & Sodium Chloride \\
Sodium Phosphate & Spironolactone \\
Sucralfate & Tacrolimus \\
Terbutaline & Theophylline \\
Ticarcillin & Ticarcillin/clavulanic acid \\
Tobramycin & Topiramate \\
Treprostinil & Trimethoprim/Sulfamethoxazole \\
Tromethamine  & Ursodiol \\
Valganciclovir & Valproic Acid \\
Vancomycin & Vasopressin \\
Vecuronium & Vitamin E \\
Vitamin K & Voriconazole \\
Warfarin Sodium
\end{tabular}
\end{center}

\end{document}